\documentclass[conference]{IEEEtran}

\usepackage{graphicx,times, amsmath, subfigure} 
\usepackage{amssymb}
\usepackage{multirow}
\usepackage[vlined,linesnumbered]{algorithm2e}
\usepackage{url}

\IEEEoverridecommandlockouts    

\begin{document}

\title{\ \\ \LARGE\bf Evolution of Things
}

\author{A.E. Eiben$^1$, N. Ferreira$^1$, M. Schut$^1$, S. Kernbach$^2$\\
$^1$Vrije Universiteit Amsterdam, \{gusz, nivea, schut\} @cs.vu.nl, \\
$^2$Universit\"{a}t Stuttgart, korniesi@informatik.uni-stuttgart.de
}

\maketitle
\begin{abstract}
Evolution is one of the major omnipresent powers in the universe that has been studied for about two centuries. Recent scientific and technical  developments make it possible to make the transition from passively understanding to actively mastering evolution. As of today, the only area where human experimenters can design and manipulate evolutionary processes in full is that of Evolutionary Computing, where evolutionary processes are carried out in a digital space, inside computers, in simulation. We argue that in the near future it will be possible to move evolutionary computing outside such imaginary spaces and make it physically embodied. In other words, we envision the ``Evolution of Things'', rather than just the evolution of code, leading to a new field of Embodied Artificial Evolution (EAE). The main objective of the present paper is to offer an umbrella term and vision in order to aid the development of this high potential research area. To this end, we introduce the notion of EAE, discuss a few examples and applications, and elaborate on the expected benefits as well as the grand challenges this developing field will have to address. 
\end{abstract}

\section{Introduction}
\label{sect:intro}
This article is a position paper about what we call embodied artificial evolution (EAE). The essence of our vision can be briefly summarized as follows. Evolutionary computing as we know it today is disembodied, performed in a digital computer space. However, recent advances (e.g., 3D printing, soft robotics, molecular engineering, synthetic biology, combinatorial chemistry, programmable matter, etc.) make it possible to move evolutionary computing out of the digital space and make it embodied. We believe that this leads to a high potential research and application area that offers great opportunities and poses great challenges. However, to realize the vision, very diverse and presently segregated fields need to interact and cross-fertilize each other. This necessitates a unifying view, an umbrella term and vision to catalyze developments in this direction. This is exactly the main objective of this paper.   

Evolutionary computing has long been used as a successful approach for solving optimisation, design, and modelling problems \cite{dejong2006evolutionary-co,Eiben2003Introduction-to}. In this context, evolutionary algorithms are traditionally locked in inside a computer. The individuals that make up the evolving population are digital objects, for example bitstrings, LISP expressions, or artificial neural nets, and all evolutionary operators (reproduction, selection, fitness evaluation) are executed inside the computer. Furthermore, evolution is used as an off-line optimisation or design tool. Even if the object to be optimised or designed is a physical one, it is only created after the evolutionary process terminated. Think of Schwefel's classic example of an optimally shaped two phase flashing nozzle or a vibration resistant satellite boom \cite{KB96}.

In this paper we argue that it is conceivable to `liberate' evolution, so that it will take place in the physical world, outside the computer, acting on tangible objects. 

The general concept of embodied artificial evolution (EAE) as assumed here can be defined by the following properties. 
\begin{enumerate}
    \item It involves physical units instead of a just group of virtual individuals in a computer.
    \item It has real `birth' and `death', where reproduction creates new (physical) objects, and survivor selection effectively eliminates them.
    \item Reproduction and selection are not executed through a centrally orchestrated main loop, but in a fully asynchronous and autonomous manner by the individuals themselves. Consequently, the population size may increase or decrease by itself. 
    \item Evolution can be driven by a combination of task-based and open-ended, environmental fitness.
\end{enumerate} 
 
\section{Discussion of the main notion}
\label{sec:notion}
To aid further elaboration about EAE systems, we consider a number of concrete examples/tasks and use these to illuminate some important aspects of EAE systems. 
\begin{enumerate}
    \item The (evolutionary) design of a robot controller for a given robot body and some task(s) in a certain environment. \\
Here, the objects to be evolved are digital, but are inherently part of a (mechatronic) physical entity. To solve this design problem one could port all evolutionary operators to the robot and execute on-the-fly evolution of controllers. Birth and death, i.e., reproduction and survivor selection, is restricted to the digital space of all possible controllers, on the robot's processors. However, fitness evaluation happens {\it in vivo} here as the reproductive probabilities of any given controller are determined by the real-world performance of the robot driven by that controller. In the present literature this approach is called embodied evolution \cite{wat02:emb} or on-line, on-board evolution \cite{Eiben2010Embodied-On-lin}.

    \item  The (evolutionary) design of a robot body for some task(s) in a certain environment.\footnote{For the sake of simplicity, let us disregard the design of the corresponding robot controller.} \\
Here, the objects to be evolved are physical. Thus, one could solve this problem by truly embodied evolution, with physical birth and death. In such a system all evolutionary operators work {\it in vivo}, including reproduction that creates new robots and survivor selection that effectively eliminates them. The main challenge here is obviously formed by the reproduction operators crossover and mutation: how to engineer a system where robots can be born (and die)? 

    \item  The (evolutionary) design of a bacterium for some medical or chemical task(s) in a certain environment. \\
Here again, the objects to be evolved are physical. However, while (re)production of mechatronic bodies is a huge challenge, bacteria reproduce by themselves. Thus, that part of the evolutionary machinery is for free in this context. The challenge here is to implement fitness evaluation and the selection operators suited to the given application objectives. Furthermore, one could implement special reproduction operators (mutation and/or crossover) that do not exist in nature, but are useful to solve the given problem.
   \end{enumerate}

We can note a couple of things about these examples that help understand some essential aspects of EAE. To begin with, observe that Example 1 is different from Examples 2 and 3 in that it is not truly embodied. To be specific, Examples 2 and 3 illustrate applications where the objects to be evolved are physical. In contrast, the objects to be evolved in Example 1 are digital, only embodied in the sense that they are hosted by a physical robot. Ironically, the term embodied evolution has been introduced for systems like the one in Example 1, cf. \cite{wat02:emb}. If needed, we can make a distinction by calling this type of systems weakly embodied and using the term strongly embodied for the ones in Examples 2 and 3. 

Furthermore, let us note that in case of a robotic application it is possible to separate the body, i.e., the physical robot with its wheels, sensors, etc. and the mind, i.e., the controller regulating the behavior of the robot. Consequently, the task of designing them also can be split into two (and combined, if needed). For the task of designing bacteria, this is not possible, because the regulatory and control mechanisms in bio-chemical organisms are not separated so clearly from the bodies to be regulated. 

Yet another difference between a robotic application and a bio-chemical one is the fact that a robotic object is more controllable for the experimenter. Robot bodies are built and robot controllers are programmed by the human experimenters. Even if we consider evolutionary development of robot bodies and controllers, the process is driven by human designed operators. These operators are usually simple; complexity emerges by their interactions. This is not the case for bio-chemical organisms, where the operators are those invented by nature. These are often very complex to understand and to manipulate. For instance, replacing one mutation operator by another one can be easy in an evolutionary robotics application, but switching off one molecular interaction and switching on another one in a cell can be (nearly) impossible.

\section{Motivations, expected benefits}
\begin{figure}[h!bt]
\begin{centering}
  \includegraphics[width=0.90\linewidth]{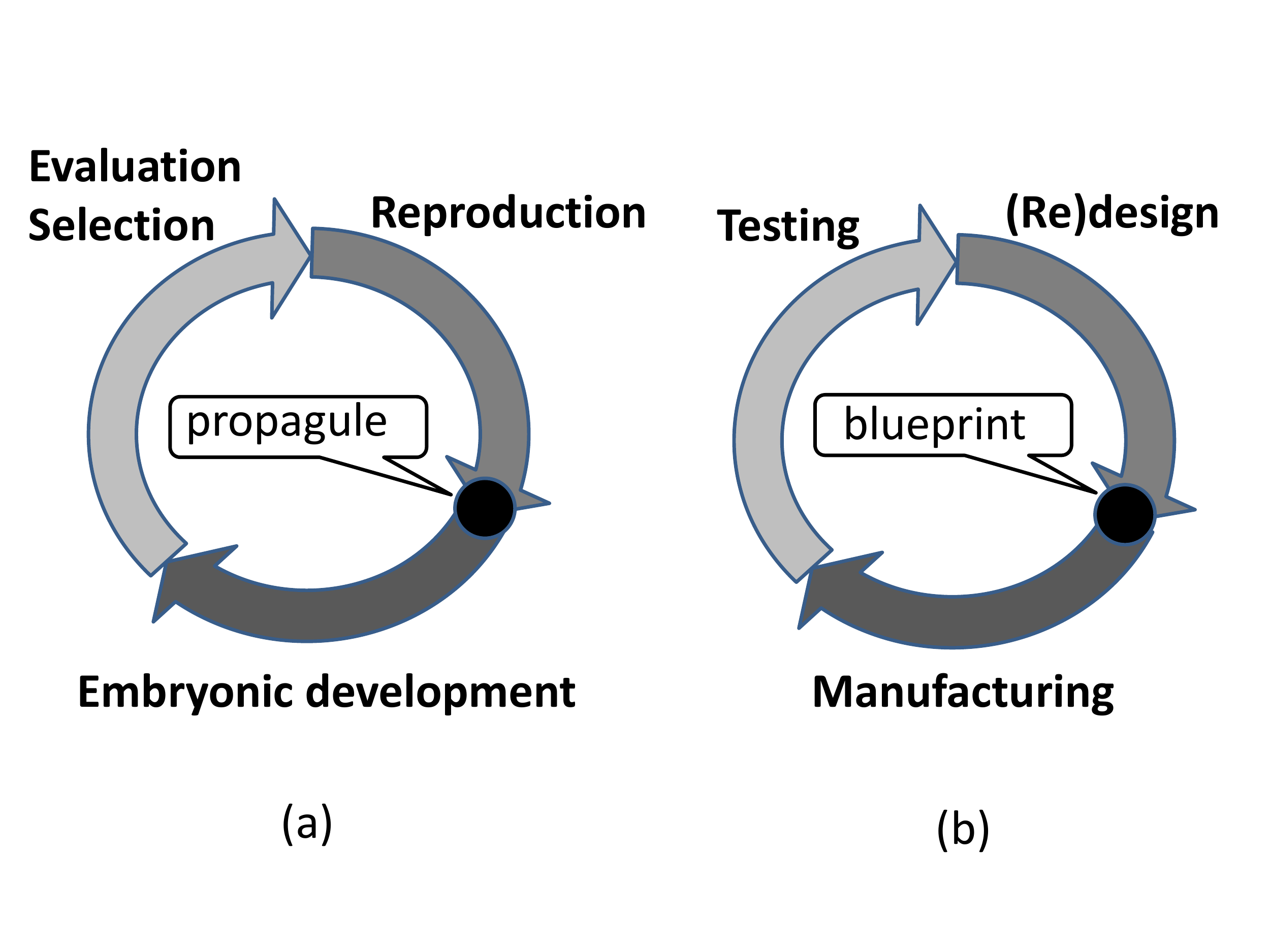}
  \caption{Two circles showing the analogies between the biological circle of reproduction (a) and the new kind of {\it in vivo} evolutionary design (b). The effective lifetime is captured by the light gray arrow labeled ``Evaluation, Selection'' and ``Testing'', respectively.}
  \label{fig:circles}
\end{centering}
\end{figure}

There are multiple reasons to investigate EAE systems.  First, EAE can lead to solving new design and engineering problems, and solving existing ones in new ways. In fact, EAE technology can be the basis of a paradigm change in how design tasks are solved. Traditionally, the design process of some artifact ends with manufacturing it. Using embodied artificial evolution, design and manufacturing become an intertwined, continuous, on-line activity, propelled by the evolutionary operators (see Figure \ref{fig:circles}). 

\vspace{-0.3cm}

Second, there is much evidence in traditional evolutionary computing showing that evolution can solve problems not solvable otherwise and that evolution can generate unexpected solutions. (Which, then, can be analysed and reverse-engineered, and thus lead to new insights and better understanding of the problem.) Once we equip certain groups of artifacts with the ability to evolve, we create the possibility that some of the evolved designs be truly original, stepping out of the box w.r.t. human thinking. 

Third, EAE systems can form the basis of a new experimentalism in biology, where evolution can be studied in a radically new way, based on controlled and repeatable experiments in a new medium. This will enable a deeper understanding of evolution in general, not restricted to (by) evolution-as-we-know-it within the existing life on Earth. 

Finally, EAE systems represent an interesting intellectual challenge. The science/art of designing and analysing evolutionary algorithms needs to be reinvented, once we change the medium from purely digital to embodied, physical. For example, population size management is trivial in a genetic algorithm, but preventing an evolving population of robots or bacteria from extinction as well as from explosion can be a hard nut to crack \cite{WSE07}. Furthermore, EAEs mean a great paradigm shift from evolving digital code to \textit{evolving things} in the real world. This implies that the environment where evolution takes place becomes orders of magnitude more complex with inherent randomness (cf. ``the noise is for free'') and a dynamics never encountered in traditional evolutionary computation.

In order to realise these and other unforeseen potential benefits, a lot needs to be done. In the remainder of this paper we elaborate on various related issues. In Section \ref{sect:areas} we provide an overview of the main areas of research on embodied artificial evolution and in Section \ref{sect:applications} we present three potential applications of EAE technology. In Section \ref{sect:challenges} we elaborate on some of the grand challenges, goals that might serve as catalysts for necessary developments and as stimulus for the specification of research agendas and milestones. Finally, in Section \ref{sect:outro} we give our concluding remarks. 

\section{Relevant Research Areas}
\label{sect:areas}
As stated previously, the realisation of embodied artificial evolution systems depends on a number of elements that are still missing. 
In this section, we describe different research areas that we consider relevant for EAE. For each area, we describe the current state-of-the-art research. We distinguish hard (i.~e., mechatrono-robotic) and wet (mainly chemistry and biology) research areas. In fact, systems can be mechatronic, bio-synthetic, bottom-up chemical, or even a hybrid of these technologies. Generally, hierarchical top-down and bottom up approaches inspired by system biology and chemistry are clearly observable; we encounter also multiple examples, such as tissue engineering, neuroscience or reconfigurable hardware, whose complexity represents a horizontal phenomenon from the viewpoint of hierarchies.

\subsection{Hardware: Mechatrono-Robotic Systems}
\label{subsect:mech}
We understand {\em mechatronics} as the synergistic combination of mechanical engineering, electronic control and information technology in order to design and manufacture useful devices. ``It relates to the design of systems, devices and products aimed at achieving an optimal balance between basic mechanical structure and its overall control." \cite{mec:els}.

According to Isermann, an intelligent mechatronic system can be developed and have the ability to model, reason and learn a process (and its functions) in order to achieve a certain goal within a given (time) frame. ``An intelligent mechatronic system adapts the controller to the mostly nonlinear behavior (adaptation),
and stores its controller parameters in dependence on the position and load (learning), supervises all relevant
elements, and performs a fault diagnosis (supervision) to request maintenance or, if a failure occurs, to
request a fail safe action (decisions on actions)." \cite{isermann02mech}.

When addressing the challenges associated with the design of mechatronic systems, Alvarez Cabrera {\em et al.} \cite{alv10:tow} state that ``The challenges are mostly related to integration of design and analysis tools, and automation of current design practices. Addressing these challenges enables the adoption of a concurrent development approach in which the synergetic effects that characterise mechatronic systems are taken into account during design." 

Applications of mechatronic systems, primarily in robotics, strongly involve the concept of embodiment~\cite{Pfeifer06:EmbodiedAI}, i.~e., using specific properties of materials to achieve a desired functionality. Examples are the locomotion for small jumping robots~\cite{Kovac08:jumping} or embodied sensor-actor coupling~\cite{Kornienko_S06}. More generally, modern robotics utilises different fields of material science, e.g.~\cite{Gong:2005p207}, which vary from modifications of surface properties up to composite materials with specific mechanical features. We have to mention here also a further miniaturisation of micro-systems~\cite{Nelson08} and structuring of material by micro-/nano- manipulation~\cite{Fatikow08}. 

In the literature various references can be found to work related to EAE in computational (mechatrono-robotic) systems. 
Watson {\em et al.} in \cite{wat02:emb,fic99:emb} envisioned embodied evolution, somewhat like we do: 
a population of individuals (in this case, robots) evolves in a completely autonomous manner.
%
Schut {\em et al} \cite{sch09:sit} present a somewhat related concept called \emph{situated evolution}, where reproduction creates new minds that become active in a pre-existing robot body, replacing an old one. 
%
In \cite{usu03:sit}, Usui and Arita address embodied evolution as in Watson {\em et al.}: robots evolve based on interactions with the environment and other robots.  
However, besides transmitting genes when robot encounters take place, 
each robot executes a genetic algorithm in order evolve a population of ``virtual individuals". 
This island model intends to speed up evolution, reducing the influence the number of real robots and the frequency in which robots encounter one another.
%
Nakai and Arita \cite{nak10:fram} extend this framework by introducing a pre-evaluation mechanism, intended to restrain robot behaviours that are estimated to be have a low fitness contribution.
%
Following then same argumentation, Elfwing {\em et al.} in \cite{elf08:bio} also make use of a subpopulation of virtual agents for each (physical) robot in order to overcome the restriction on population size. 

The following examples take the definition of EAE a step further than the previous ones: evolutionary operators and principles are not to be applied to robot controllers only but to the robots themselves. 
%
Firstly, Pollack {\em et al.} present in \cite{pol01:three} 
systems working towards a fully automated manufacturing of autonomous robots. Co-evolution of mind and body of robots is their central issue. 
The most advanced type of robots are modular robots, where a set of rules that generate the robot structure is evolved. 
%
Secondly, another example of modular robot systems is presented in \cite{zyk07:evolv}, where Zykov {\em et al.} discuss self-replication:
 ``composed from robotic modules that can change their shape and topological connectivity", making the overall morphology of a robot adaptable. The replication process starts when a parent creates a new unit, which in its own turn can also assume the role of parent and create other copies of itself. 


\subsection{Wetware: Bio-Chemical Systems}
\label{subsect:wetware}
Biological systems have a clear advantage over mechatronic devices, as biological properties, such as reproduction, can be taken for granted. For instance, a biological system is naturally equipped to carry out evolutionary processes. Not only reproduction and self-preservation, but also selection and adaptation capabilities are inherent to the system. However, an important challenge is how one can manipulate the system in order to obtain exactly what one is looking for. Or, in other words, how to {\em program} a bio-synthetic system? In Section \ref{ssubsect:bio} we given some examples on how this has been accomplished so far.

The bio-synthetic systems utilise existing biological cellular systems with their very complex metabolism. The approach from bottom-up chemistry uses another methodology: creating elementary basic cellular (so-called vesicles) and multi-cellular structures ``from scratch". Advantages of this approach are multiple degrees of freedom in designing metabolic networks (in simple cases -- autocatalytic reactions) and different internal and external interaction mechanisms. Example of such chemical systems are presented in Section \ref{ssubsect:chemo}.

\subsubsection{Top-down Biological Systems}
\label{ssubsect:bio}
Programming cells does not have the purpose of substituting silicon computing, but being able to have access to the numerous functionalities and properties on those cells in a predictable, reliable way. Obtaining such programmable bio-devices, or embodied information processors, is a major issue. Potential applications include three-dimensional tissue engineering, biosensing and biomaterial fabrication.

Natural processes can be often described in terms of a networks of simple computational components, or {\em biobricks} \cite{amos09:bac}. When referring to biological computing, the main objective is to use the power of natural processes for the purpose of computation. For instance, attempting to ``reprogram" parts of a bacterial cell in order to make it perform a certain task. Because natural processes are intrinsically random, changing functionalities of a cell, as well as adding new desired behaviours, is not a trivial exercise. Differently from mechatronic systems, where circuits are built with an specific {\em known} purpose, the circuitry of a cell may not be completely understood.

However, advances in the area of synthetic biology have allowed some interesting recent results. For instance, in \cite{tam10:rob}, Tamsir {\em et al} show how logic gates can be built in {\em Escherichia coli} cells and how complex computations can be produced by ``rewiring" communication between cells. ``Analogous to a series of electrical gates arrayed on a circuit board, compartmentalisation of genetic gates in individual cells allows them to be added, removed or replaced simply by changing the spatial arrangement of the E. coli strains." \cite{tam10:rob}.

Using an alternative approach, Rigot {\em et al} describe in \cite{rig10:dis} how to implement complex Boolean logic computations, which reduces wiring constraints. This is obtained through a redundant distribution of the desired output among the engineered cells. In detail, the engineered yeast cell sense its surroundings (following certain criteria) and send signals to other cells (via secreted wiring molecules). Following the idea of biobricks, a number of yeast cells can then be combined into more complex circuits.

\subsubsection{Bottom-up Chemical Systems}
\label{ssubsect:chemo}
Examples of bottom-up chemical systems can be found in artificial chemistries~\cite{Dittrich01}, self-replicating systems~\cite{Hutton09}, using bio-chemical mechanisms for, for example, cognition~\cite{Dale10}. In several works, this approach is denoted as swarm chemistry~\cite{Sayama09}. Researchers hope that such systems will give answers to questions related to developmental models~\cite{Astor00}, chemical computation~\cite{Berry92}, self-assembly, self-replication, and simple chemistry-based ecologies~\cite{Breyer97} or technological capabilities of creating large-scale functional patterns \cite{Yin08}. Several approaches consider meso- and nano-objects, such as particles with functionalised surfaces~\cite{Schmid04}, colloidal systems~\cite{Fujita09}, or molecular networks~\cite{Nitschke09}; a system of elementary autonomous agents, which possess rudimentary capabilities of sensing and actuation. Information processing and collective actuation is performed collectively as, for example, stochastic behavioural rules. Several phenomena, such as meso-scale self-assembling or diverse self-organising processes~\cite{Davies09}, make these type of systems attractive in applications. Projects such as ECCell, BACTOCOM, NEUNEU, MATCHIT are addressing the questions of chemo-ICT interfaces.

Molecular, colloidal and particle systems also use local interactions and horizontal mechanisms, similarly to 2D and 3D ecological swarms, however we can observe another approach for designing collective phenomena, using the same very simple but large-scale interaction patterns for whole systems~\cite{Kumar06}. Many research projects in collective nanorobotics and molecular systems are focusing on the technological capabilities of creating such large-scale patterns, e.g.,~\cite{Yin08}.

\subsection{Hybrid Mechatronic and Biochemical Systems}
\label{subsect:hybridSystems}
Examples of hybrid mechatronic and biochemical systems are bio-chemical and microbiological systems: using bacterial cellular mechanisms~\cite{Wood99} as sensors, the development of bacterial bio-hybrid materials~\cite{Ruiz-Hitzky10}, the molecular synthesis of polymers~\cite{Pasparakis10} and biofuels~\cite{Alper09}, genome engineering~\cite{Carr09}, and more general fields and challenges of synthetic biology~\cite{Alterovitz09}. Example of hybrid technologies are attempts to interact with biological populations by means of technological artifacts, for example to manage the grazing of cattle over large areas~\cite{SchwagerJFR08}, \cite{correll:SAB2008}, to control mixed societies of robot and insects~\cite{caprari05}, or to encourage social communication between robots and chickens~\cite{Gribovskiy09}. A similar approach is related to the integration of different robot technologies into human societies, for example the management of urban hygiene based on a network of autonomous and cooperating robots~\cite{Mazzolai08}.

\section{Applications}
\label{sect:applications}
The proof of the pudding is the eating. A new technology is largely justified by useful applications. In the present embryonic stage of the EAE field, it is impossible to predict what the best applications will be. However, it is possible to identify some application domains that can demonstrate the benefits of EAE systems in the short/mid-term. 

In general, the problems addressed by EAE systems should have a high level of difficulty, including (but not restricted to): (i) changing environments; (ii) multitasking and multi-objective applications; (iii) problems where robust solutions are required; and, (iv) on designing systems with emergent (self-organising) behaviour. Furthermore, empowered by evolution, systems might present other interesting characteristics such as the capability of learning about and adapting to the environment. Additionally, depending on the application at hand, features such as personalisation of its components, where the machinery becomes adequate to one's personal needs and requirements, may also be of interest. 
  
\subsection{Robot Companions} 
One could imagine a whole ecosystem of robot companions or artificial pets. Regarding their bodies, these could range from a few cubic decimeters (cat and dog size, if you wish) up to human comparable sizes. As for their mental features, they should be gentle, caring, helpful, and to some extent even intelligent. From a functional point of view they could perform specific tasks, (simple domestic tasks, health monitoring, alarm) and/or provide more generic `emotional' services (keeping company, being good listeners, acting as partners in simple conversations). Embodied artificial evolution can be a key enabler here, where breeding is the obvious natural analogy. (Note, that breeding as we know it for thousands of years, can bee seen a special case of EAE, where mating selection is artificial, while survivor selection and reproduction are natural operators.) As indicated by the examples in Section \ref{sec:notion}, design and implementation of reproduction operators forms a major technical challenge here. Material science is a key discipline to this end.  

\subsection{Environment-friendly Organisms}
In April 2010 the largest oil spill in US history happened: the equivalent of around 4 million barrels of oil flowed into the Gulf of Mexico, with numerous ecological implications. Analysis on the site, a couple of weeks after the disaster, showed that many groups of bacteria were helping to clean up the waters. These bacteria were able to break down the chemicals found in crude oil and, in fact, responded quite effectively to the incident. However, given the scale of the spill and the fact that some components of the oil could not be broken down by these organisms, the performance of the biological ocean crew was limited. One can imagine a scenario in which such bacteria are synthetically designed. This could have two clear advantages: (i) a large number of organisms could then be generated and deployed, in order to correspond to the scale of the disaster; and, (ii) the bacteria's digestive capabilities could be adjusted in order to guarantee that all oil components could be consumed. In a matter of hours, instead of weeks or months, the waters would be clean, preventing any mid- or long-term consequences for the environment. Such artificially developed organisms can also be used to create building material and biofuel, data storage and stopping desertification.

\subsection{Evolutionary 3-D Printing}
Imagine a world in which anything can be produced with just a few clicks. Customised products are at the reach of your hand, ranging from a child's toy to a meal. Vilbrandt {\em et al.} introduce in \cite{vil08:univ} the {\em universal desktop fabrication} (UDF) that can produce essentially any complete, finished, and functional object. The Fab$@$Home (www.fabathome.org) is a desktop rapid prototyper and a first step towards UDF. Such personal fabricators can build objects from different materials and will allow anyone to build functional objects that fit its own needs and desires. ``You can imagine a 3-D printer making homemade apple pie without the need for farming the apples, fertilising, transporting, refrigerating, packaging, fabricating, cooking, serving and the need for all of the materials in these processes like cars, trucks, pans, coolers, etc,"\footnote{Homaro Cantu states in the BBC News article ``The printed future of Christmas dinner": http://www.bbc.co.uk/news/technology-12069495}. Additionally, you can think of future fabricators that are able to check what is being printed and adjust the design and mixture of materials in order to obtain the best results. 
(Embodied) Evolution is expected to play an important role in the development of such fabricators, cf. \cite{rie06evofab,rie10:evo} : ``Ultimately, the evolution of form and formation become fully intertwined when the language of assembly itself becomes subject to evolution [\ldots]. Through this co-evolution of form and formation, Evolutionary Fabrication discovers both how to build objects and what to build them out of.'' 

\section{Grand Challenges}
\label{sect:challenges}
At this moment it is impossible to foresee how this field will develop. However, even in the present early stage, we are able to identify some of the grand challenges that certainly will have to be addressed. In this section we discuss five of these.

\subsection{Body types}
The essence of embodied evolution is -- the body. To this end, we can distinguish mechatrono-robotic systems (hardware) and bio-chemical systems (wetware), that may also be hybridised. Regarding wetware, there are two options again: bottom-up, relying on chemistry, or top-down, based on biology.  Recent developments in microfluidics, functional fluids, or programmable matter also seem very promising. The first grand challenge is thus to find body types suited for (self-)~reproduction.
Here we can identify several fundamental questions, the most important of those being: \emph{How can the current ICT be combined with bio-chemical developments?} This question is also known in other formulations, as e.g. ``programmability of synthetic systems", or ``open-ended embodied evolution", and is one of the key points in understanding principles of synthetic life. It is also addressed by the European bio-ICT initiative and several research projects, e.g. PACE~\cite{PACE} and e-FLUX~\cite{e-Flux}, just to name a couple. 

\subsection{How to Start --  Robot Reproduction}
 The implementation of birth (reproduction operators) for human engineered physical devices is a critical prerequisite for EAE. These operators must also realise some form of inheritance. The three approaches based on mechatronics, chemistry, or biology differ greatly in this respect. (Self-)reproducing mechantronical and chemical units are far from being trivial, whereas it comes for free in biological systems.

In current mechatrono-robotic systems there are two concepts that are crucial for EAE: self-assembling and self-replication. Self-assembling is a process which creates complex systems from basic elements, whereas self-replication means a reproduction of these basic elements. The self-assembling has already been targeted several times in robotics, e.g.~\cite{MiyashitaHH07},~\cite{KernbachDARS10}. The field of modular (reconfigurable) robotics aims at self-assembling of robot modules into complex structures, so called artificial organisms~\cite{Levi10}. Robots are able to make functional copies of artificial organisms from basic building blocks provided there exists an essential reserve of such basic modules.

The self-replication of basic elements remains so far unsolved, even in principle. The problem lies in a high technological complexity of functional units, such as motors, gears or microelectronics. There are several attempts to address self-reproduction in modular robotics, e.g.~\cite{zyk07:evolv}, by additive plastic molding~\cite{sells_reprap:replicating_2009} (see also RepRap.org), or by using 3D prototyping technology~\cite{rie10:evo}. No one of these technologies apply self-reproduction as introduced above. For that, further developments are still necessary. This relates not only to the creation of physical units, but also to the computing capabilities, as well as inheritance mechanisms.

\subsection{How to Stop -- Kill Switch}
A very serious concern regarding EAE is the possibility of `runaway evolution'. By this term we do not mean the Fisherian notion of sexual selection reinforcing useless traits \cite{r.a.fisher1930the-genetical-t}. Runaway evolution as we use it here stands for the process of uncontrolled population growth. Such a growth might also be accompanied by the emergence of new, unwanted features in the population. Obviously, it would be highly irresponsible to expose ourselves to such a risk. To reduce this risk, all such experiments could be carried out in highly secured isolated environments, not unlike current research into certain germs, bacteria, viruses, etc. However, this might disable the whole application in cases where the evolving population is inherently free, where the individuals are part of our everyday life (robot companions, waste-eating organisms, medical nano-robots in the human body, etc.). In such cases a `kill switch' is required to guarantee that human supervisors are able to shut down the system, if and when they deem necessary.  

\subsection{Evolvability and rate of evolution}
It is well-known in biology as well as in evolutionary computing that evolution takes time. One might even say that evolution is slow. Although this statement obviously depends on the context dependent notion of time, it is safe to say that, in general, it takes many generations to achieve a decent level of development. Useful EAE systems must exhibit a high degree of evolvability and a high rate of evolution \cite{Hu2010Evolvability-an}. In practice, they must achieve decent progress in real time: have short reproduction cycles and/or large improvements per generation. The main factors here are the application dependent time requirements and the speed of progress, determined by the evolutionary operators. For instance, medical nano-robots put at work in a human body should adapt within a few hours to their environment (the patient's body). In case of sending evolving robot explorers with a rough initial design to Mars, one can wait months for appropriate designs to emerge. 

Building fast evolutionary systems is a nontrivial challenge on its own. Failing to meet this challenge would imply that the real time performance of EAE systems is too low. Ultimately, this could even disqualify the whole approach -- at least, for certain applications. In general, the speed of evolution should be used as one of the essential assessment criteria for judging the feasibility of potential applications. 

\subsection{Process control \& Methodology} A radical change caused by EAE technology is that design and manufacturing become an intertwined, continuous activity. On the one hand, this allows systems to be autonomous and self-improving. On the other hand, this poses an unprecedented challenge for maintaining human control during the process. In particular, human users should be able to perform on-line monitoring and steering in line with the given user preferences. Technically this means directed evolution that could be perhaps realised by directed selection (akin to breeding) and/or directed reproduction (as in genetic manipulation). On a conceptual level, this requires a new kind of methodology that must contain traditional elements, such as specification and validation as well as address previously unforeseen aspects, e.g.,  mixing (the dynamics of) ``free'' evolution with specific design objectives on-the-fly. Part of this challenge  --or perhaps a challenge on its own-- is the `freeze switch', that is, the ability to recognize if/when the evolving objects have obtained the required properties and stop further evolution without killing the system. 

\section{Final remarks}
\label{sect:outro}

In this paper we have presented the concept of Embodied Artificial Evolution or the Evolution of Things. The systems we envision here are indeed embodied and artificial. They are embodied because evolutionary operators (reproduction, selection, fitness evaluation) are implemented in/by the physical objects that undergo evolution. Furthermore, they are artificial because (i) the individuals and the population as a whole are being designed and/or programmed to fulfill a certain human purpose, to execute a certain task (not excluding open-ended evolution to take place in parallel), and (ii) the evolutionary operators and their particular combination into one working system are human engineered. 

We have briefly discussed certain areas of research that can be the basis for developing the physical grounding of such systems. We arranged these areas in a simple taxonomy consisting of hardware (mechatrono-robotic approach), wetware (bio-chemical systems), and their possible hybrid and mentioned a few emerging areas, such as microfluidics, functional fluids, and programmable matter. We have also provided some potential application examples of EAE and have discussed some of the grand challenges lying ahead, from physical reproduction of robots to monitoring and controlling a running EAE application.

In conclusion, we believe that Embodied Artificial Evolution forms a high potential research and application area. This field is in an embryonic stage, where relevant developments take place within different scientific fields and technological areas that do not naturally interact with each other. We hope that by introducing an umbrella term and a unifying vision we can help bring them together through raising awareness of the shared research issues and possible solutions. As of today, some elements of EAE systems already exist, but considerable scientific and technological advances are necessary to achieve the vision sketched here. However, we do expect that the first examples of such systems will arise in the near future.

\section*{Acknowledgments}
We gratefully acknowledge the financial support of the European Commission funding the EvoBody project (grant number FP7-258334), as well as the contribution of the experts we have consulted during the project, cf.~http://www.evobody.eu/. 

\bibliographystyle{plain}
\newcommand{\Sortkey}[1]{}

\end{document}